# HamRaz: A Culture-Based Persian Conversation Dataset for Person-Centered Therapy Using LLM Agents

**Mohammad Amin Abbasi[1], Farnaz Sadat Mirnezami[2], Ali Neshati[2], Hassan Naderi[1*]**

[1]Department of Computer Engineering, Iran University of Science and Technology, Tehran, Iran
[2]Computer Science, Faculty of Science Ontario Tech University Ontario, Canada

m_abbasi1378@comp.iust.ac.ir
farnazsadat.mirnezami@ontariotechu.net
naderi@iust.ac.ir
ali.neshati@ontariotechu.ca

## Abstract

We present HamRaz, a culturally adapted Persian-language dataset for AI-assisted mental health support, grounded in Person-Centered Therapy (PCT). To reflect real-world therapeutic challenges, we combine script-based dialogue with adaptive large language models (LLM) role-playing, capturing the ambiguity and emotional nuance of Persian-speaking clients. We introduce HamRazEval, a dual-framework for assessing conversational and therapeutic quality using General Metrics and specialized psychological relationship measures. Human evaluations show HamRaz outperforms existing baselines in empathy, coherence, and realism. This resource contributes to the Digital Humanities by bridging language, culture, and mental health in underrepresented communities.

## 1 Introduction

Recent advancements in Large Language Models have significantly expanded the potential of AI in supporting mental health, particularly through simulating therapeutic conversations. These models are capable of producing coherent, context-sensitive, and emotionally resonant responses, making them increasingly useful in preliminary mental health screenings, emotional support systems, and AI-powered conversational agents (Hua et al., 2024; Stade et al., 2024). Despite these achievements, most research in this domain remains focused on English and East Asian contexts, overlooking the cultural and linguistic diversity required for truly inclusive mental health support.

In particular, Persian-speaking communities remain underserved in this space. Cultural norms, language structures, and societal values deeply influence the therapeutic process, affecting both how individuals express emotional distress and how they respond to interventions. Yet, no comprehensive efforts have been made to build culturally relevant, Persian-language datasets or therapeutic AI systems, leaving a significant gap in both research and application.

Moreover, while Cognitive Behavioral Therapy (CBT) (Beck, 1979) has commonly been adopted in AI-based simulations, its dependency on diagnostic precision and structured intervention frameworks poses a challenge for LLMs, which lack clinical reasoning capabilities. In contrast, Person-Centered Therapy (PCT) (Rogers, 1951)—a humanistic approach that emphasizes empathy, non-directiveness, and the client's own voice—aligns more naturally with the conversational and generative strengths of LLMs, making it a more suitable foundation for AI-driven therapeutic dialogues.

Existing approaches to generating psychotherapy datasets, such as Two-Agent Mode (fully LLM-driven simulations) (Zhou et al., 2023) or Script-Based methods (predefined conversation templates), often fall short in realism and complexity. These methods tend to produce unnatural or overly simplistic dialogues, failing to capture the ambiguity, emotional conflict, and indirect expression typical of real therapeutic settings. They also frequently portray therapy as overly effective within a single session, misrepresenting the gradual and nonlinear nature of psychological healing.

To address these issues, we introduce HamRaz, the first Persian-language dataset designed for PCT-based AI therapy. Our approach combines the structural clarity of scripted scenarios with the adaptability of LLM role-playing to generate more realistic and emotionally rich interactions. In addition, we propose HamRazEval, a two-tier evaluation framework that measures both general conversational quality and therapeutic depth using adapted Barrett-Lennard Relationship Inventory metrics (S. Chen, Liao, F., Murphy, D., & Joseph, S. , 2023). This contribution aims to advance culturally sensitive mental health AI by providing a novel resource tailored to Persian-speaking users, while offering generalizable insights into human-aligned therapy simulations.

* Corresponding author.

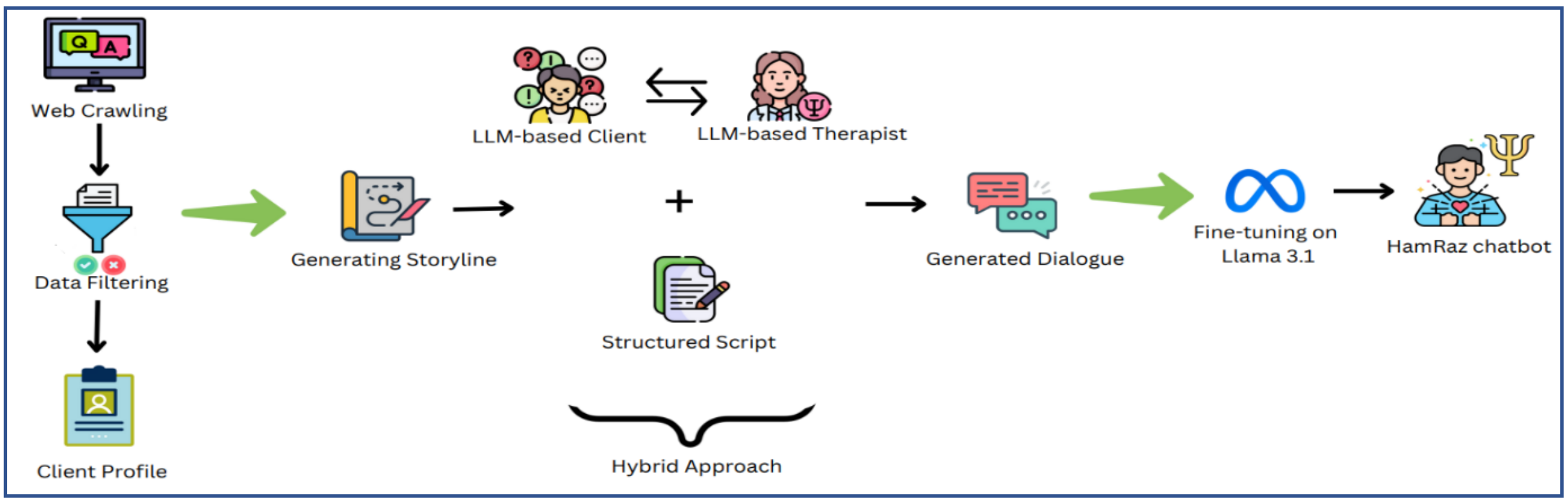

Figure 1: The architecture of the HamRaz simulation framework.

The overall architecture of the HamRaz simulation framework is illustrated in Figure 1.

## 2 Related Works

### 2.1 Dataset for Simulated Counseling and Therapy

Several studies have explored the use of large language models to simulate counselor-client interactions for generating training data in LLM-based therapy models. For instance, Psych8k (Liu et al., 2023) includes 8,187 instruction pairs from 260 counseling sessions. CPsyCounD (Zhang et al., 2024) contains 3,134 multi-turn consultation dialogues generated based on real reports in Chinese contexts. PATIENT-Ψ (Wang et al., 2024) emphasizes LLM-generated patient interactions for mental health training. Moreover, CACTUS (Lee et al., 2024) is a multi-turn dialogue dataset designed to emulate real-life counseling interactions using Cognitive Behavioral Therapy techniques.

In spite of these developments, challenges remain regarding cultural and language adaptation and the integration of Person-Centered therapeutic approaches.

### 2.2 LLMs for Mental Health Applications

The application of large language models in psychological counseling and mental health support is a growing area of research. several models have been developed with a focus on mental health interventions. Psy-LLM (Lai et al., 2023) integrates pre-trained LLMs with professional psychological content to enhance counseling responses, while ChatCounselor (Liu et al., 2023) is a fine-tuned model based on the Psych8k dataset, designed to improve mental health support. Additionally, SoulSpeak (Zhang & Luo, 2024) incorporates a dual-memory system for long-term context retention, while a hybrid model explored in (Yu & McGuinness, 2024) combines DialoGPT and ChatGPT-3.5 for psychological support. PsyChat (Qiu et al., 2023) and SimPsyBot (Qiu & Lan, 2024) simulate counselor-client interactions, enhancing the realism of therapeutic dialogues.

Despite these advancements, most models lack cultural adaptation for Persian-speaking users. Our work addresses this gap by integrating person-centered therapy principles into a Persian-language LLM-driven therapy framework, enhancing cultural relevance and accessibility.

## 3 Methodology

### 3.1 Data Collection

We developed a dataset for Persian person-centered therapy by crawling publicly accessible psychological question-and-answer interactions. The data was sourced from Iranian psychology websites and forums, including ehyacenter[2], moshaverfa , and simiaroom, where individuals from Iran openly post their mental health queries and psychologists provide answers.

[2] ehyacenter.com

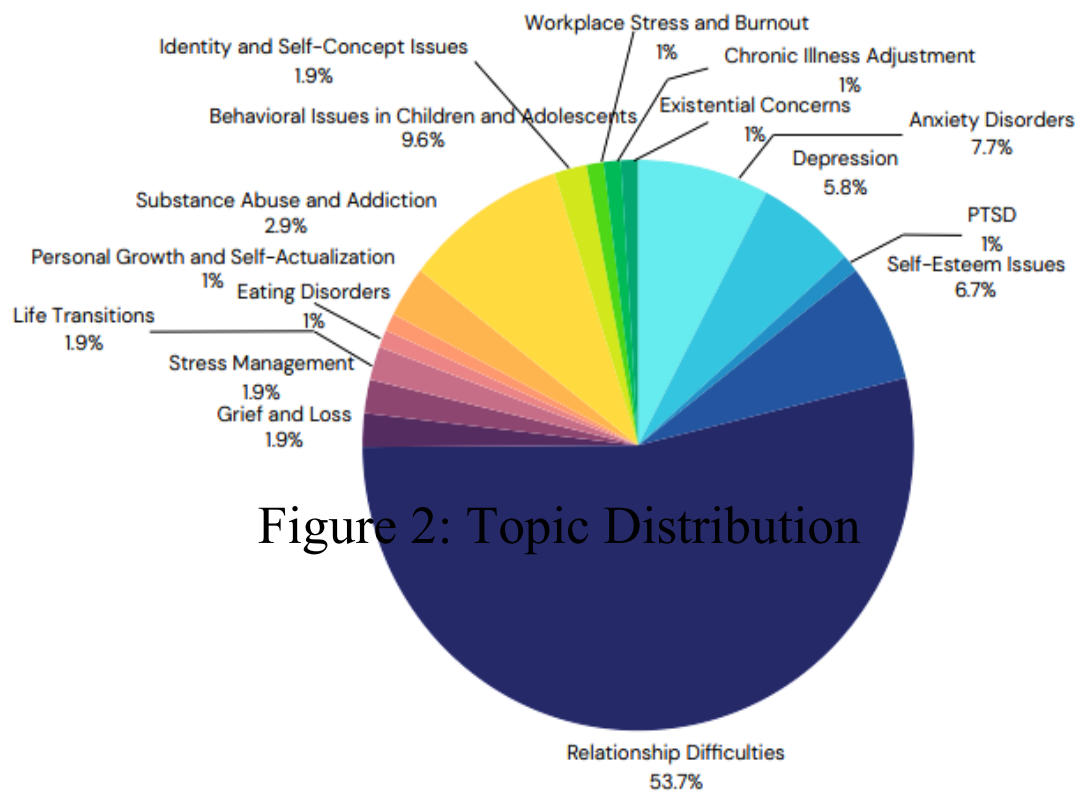

Figure 2: Topic Distribution

Given that all crawled content was already in the public domain and universally accessible, this approach ensures the dataset accurately captures real-world psychological concerns and therapeutic dialogues prevalent in the Iranian cultural setting.

Through this process, we successfully collected 4,000 user-submitted questions, covering a wide range of mental health issues. The dataset's authenticity and cultural relevance were maintained by selecting platforms frequented by Persian-speaking users seeking psychological advice. These questions and problem statements inherently reflected cultural nuances, societal expectations, and region-specific challenges, making them particularly valuable for developing a context-aware person-centered therapy model. You can find a Sample of questions in Appendix B.

### 3.2 Ensuring Data Privacy

To ensure the privacy and confidentiality of the dataset, we undertook a meticulous process of reviewing and preprocessing each text entry. This step was essential to mitigate the risk of exposing sensitive or personally identifiable information and to prepare the data for subsequent analysis in an ethical and responsible manner. We carefully anonymized all data entries to remove or obscure any information that could directly or indirectly identify individuals. Specific measures included:

- Personal names were replaced with generic labels such as "Person A" or "Participant X."
- Specific geographic details were replaced with broader terms like "residential area" or "large city.
- Exact dates were generalized to months or even years, depending on the context.
- Any unique identifiers, such as phone numbers, email addresses, or social media handles, were completely removed.
- Contextual Adjustments: Context-specific identifiers, such as job titles ("CEO of X Company") or project names, were anonymized to generic terms like "senior manager" or "project leader."

### 3.3 Topic Classification and Filtering

To refine the dataset and ensure alignment with Person-Centered Therapy (PCT) principles, we employed GPT-4o (OpenAI, 2023) with a structured prompt to classify the 4,000 collected questions into 16 predefined mental health categories. These categories were selected based on their therapeutic relevance to PCT, covering a wide range of psychological concerns such as anxiety disorders, depression, PTSD, self-esteem issues, relationship difficulties, grief and loss, stress management, life transitions, personal growth, eating disorders, substance abuse, behavioral issues in youth, identity and self-concept challenges, workplace stress, chronic illness adjustment, and existential concerns.

GPT-4o was prompted to follow a binary classification process, assessing whether each question matched one or more of the predefined topics. Non-relevant questions, such as those focused on medical or technical advice, were automatically excluded. The full prompt and topic definitions are provided in Appendix A. This process ensured that the dataset remained focused, culturally relevant, and suitable for person-centered therapy applications. The distribution of topics is shown in Figure 2.

### 3.4 Client Profiling

Once the dataset was refined to include only therapeutically relevant questions, the next step involved profiling each user submission to extract key psychological insights. This profiling process aimed to categorize emotional themes, psychological issues, and contextual factors, enabling a deeper understanding of user concerns and facilitating more nuanced therapy simulations.

To achieve this, we utilized GPT-4o to analyze each message and generate a structured profile capturing: (1) Emotional Themes (e.g., fear, sadness, insecurity), (2) Key Psychological Issues (e.g., trust problems, abandonment, low self-esteem), (3) Past Experiences (e.g., childhood trauma, family

**Client Question:**
"Hello,
I struggle with assertiveness in tasks that require interaction and collaboration with others. Sometimes, being too available, compromising too much, accepting others' mistakes, and constantly giving in make me feel exhausted and worthless. In response, I become stubborn as a way to make those around me realize how they've taken advantage of my flexibility.
I want to understand the difference between stubbornness and assertiveness. Also, I need guidance on how to recognize when I should be assertive. I naturally tend to be soft and flexible, but I feel that it's not always the right approach."

**Client Profile:**

```
{
  "emotional_themes": ["frustration", "insecurity", "exhaustion", "confusion", "desire for assertiveness"],
  "key_psychological_issues": ["lack of assertiveness", "fear of being taken advantage of", "self-worth issues", "difficulty in setting boundaries"],
  "past_experiences": ["experiences of being overly accommodating in relationships", "feelings of being undervalued or unappreciated"],
  "patterns_and_behaviors": ["over-accommodating behavior", "difficulty in asserting needs", "oscillation between flexibility and stubbornness"],
  "desired_outcome": "guidance on distinguishing between assertiveness and stubbornness, and strategies for being more assertive",
  "contextual_factors": ["the user is likely in a collaborative work environment", "the user may have a tendency to prioritize others' needs over their own"]
}
```

Figure 3: Example of a client question and its generated profile.

conflict), (4) Patterns and Behaviors (e.g., overthinking, avoidance, reassurance seeking), (5) Desired Outcomes (e.g., coping strategies, emotional relief, validation), and (6) Contextual Factors (e.g., age, family dynamics, workplace or societal stressors).

The profiling process enriched the dataset by categorizing user concerns into key psychological themes, allowing the LLM-driven therapy model to generate more tailored and empathetic responses. Using a structured prompt-based approach (detailed in Appendix A), the model analyzed, categorized, and extracted relevant insights from user messages. This method ensured that each concern was not only thematically classified but also contextually understood, enhancing the dataset's psychological depth and enabling more realistic therapy simulations. An example of a generated client profile is in Figure 3.

Across 3,400 profiled messages, multiple emotional themes were typically present in each query. The most frequent themes were frustration (2,914), sadness (2,124), anxiety (1,306), fear (1,228), insecurity (1,224), and confusion (1,150). Less frequent but notable emotions included concern (678), helplessness (415), disappointment (266), and guilt (216). This distribution highlights that relational strain, self-worth challenges, and anxiety-related struggles are among the most prominent concerns expressed by Persian-speaking clients in the dataset.

### 3.5 Adding Complexity to Client Statements

To improve the realism of client interactions, we introduced complexity in half of the user profiles by modifying their statements to reflect unclear, indirect, or conflicting expressions. This step aimed to simulate real-world therapy challenges, where clients may struggle with articulating emotions, expressing mixed feelings, or avoiding sensitive topics.

Using GPT-4o, we applied a structured complexity-enhancement prompt (detailed in Appendix A) to analyze user profiles and assign relevant characteristics to each case. These characteristics were categorized into Unclear Statements, Indirect Statements, Conflicting Statements, Mixed Emotions, Avoidant or Defensive Responses, Cultural Ambiguities. Each user's psychological profile and contextual factors were used to assign one or more complexity traits, ensuring that therapy dialogues remained nuanced and reflective of real client experiences.

### 3.6 Simulating Psychotherapy Session

We structured the simulated therapy sessions using a five-stage framework based on Person-Centered Therapy (PCT) principles. These stages were designed to mirror real-life therapy sessions, guiding the interaction between the client and psychologist in a natural, progressive manner. By systematically defining each stage, we ensured that LLM-driven therapy simulations maintained coherence, therapeutic effectiveness, and emotional realism.

The simulated sessions followed five PCT-based stages: (1) building rapport and trust, (2) active empathetic listening, (3) encouraging self-exploration, (4) supporting growth and change, and (5) reviewing and closing.

To ensure realism in AI-driven therapy simulations, we used a structured prompt-based process that mapped each client profile onto five therapy stages. At each stage, the model selected emotions, expressions, and behaviors from predefined options, creating coherent and fluid dialogues that captured emotional ambiguity and nuanced struggles, resulting in consistent and realistic therapeutic interactions. The structured prompt design is detailed in Appendix A, and an example is provided in Appendix B.

### 3.7 Generating a Storyline

We improved the clarity and authenticity of therapy dialogues by first generating a narrative storyline before converting it into dialogue. This intermediate step ensured that therapy interactions felt natural and human-like, rather than fragmented LLM-generated responses. Using GPT-4o, we structured therapy sessions into a five-stage narrative, integrating the client's emotional themes, psychological issues, and past experiences. This approach provided a logical emotional progression, aligning client interactions with realistic therapeutic dynamics.

A storyline-first approach ensured a natural emotional arc, avoiding disjointed exchanges. It also allowed for richer context, incorporating body language, pauses, tone shifts, and facial expressions—elements often missing in direct LLM-generated dialogue. Additionally, by shaping the session as a narrative before dialogue conversion, we improved control over the conversation's flow, leading to more lifelike therapy simulations in Persian. The full prompt and an example of this process are detailed in Appendix A and B.

### 3.8 Converting the Storyline into Dialogue Format

Our methodology aligns with findings from CACTUS research (Lee et al., 2024). This study compared two methods for generating psychotherapy dialogues: (1) Two-Agent Mode (Zhou et al., 2023), where separate models assume the roles of the client and therapist, and (2) Script Mode, where dialogue is generated from a structured script with predefined client and counselor information. (see Appendix A and B)

#### 3.8.1 Generating Dialogue using Script Mode

To generate structured therapy dialogues, we transformed the narrative-based therapy sessions into a scripted dialogue format between a therapist and a client. This approach ensured that conversations maintained a natural flow and therapeutic coherence, reflecting the dynamics of real-world psychotherapy sessions.

Results in CACTUS research demonstrated that scripted dialogue generation produces more natural and well-constructed conversations compared to Two-Agent Mode. Inspired by these findings, we adopted a scripted approach, ensuring that the therapist-client interactions were coherent, psychologically grounded, and contextually aligned with Person-Centered Therapy principles.

To guide this transformation, we utilized a detailed prompt-based methodology, incorporating client profiles, emotional themes, past experiences, and cognitive patterns into the dialogue generation process. This approach enabled us to create a high-quality, culturally nuanced dataset of Persian therapy conversations, allowing for more effective LLM-driven psychotherapy sessions simulations.

#### 3.8.2 Role-Playing LLM-to-LLM Interactions

We introduce a novel hybrid approach to improve the dynamic quality, conversational depth, and psychological validity of therapy dialogues. This approach integrates script mode with two-agent mode, combining the structural benefits of pre-scripted dialogues with the flexibility of agent-based interactions. It enables two LLM-based agents to simulate a therapist-client conversation, refining the dialogue dynamically while maintaining adherence to Person-Centered Therapy (PCT) principles.

In this framework, one agent assumes the role of the therapist, guided by a system prompt designed to enforce PCT techniques such as reflective listening, open-ended questioning, and non-directive engagement. The second agent assumes the role of the client, ensuring responses align with a predefined user profile, including emotional states, past experiences, and psychological patterns. Unlike conventional two-agent systems, where LLMs interact freely and often generate inconsistencies, our approach structures these interactions around a pre-scripted baseline dialogue, which is iteratively refined at each conversational turn.

#### 3.8.3 A Hybrid Approach: Combining Script Mode and Two-Agent Mode

While pre-scripted dialogues offer consistency, they often lack the spontaneity and adaptability essential for realistic therapeutic interactions, frequently failing to adjust to subtle shifts in client emotions and the natural flow of conversation. To overcome these limitations, we introduce a hybrid approach that synergizes script mode with two-agent interactions. This method utilizes pre-scripted dialogues as a foundational framework, which is then dynamically refined by LLM agents at each conversational turn, ensuring dialogues remain structured yet flexible and capable of incorporating real-time adjustments without sacrificing coherence.

Our initial scripted dialogues, though consistent, tended to be mechanical and lacked the depth and natural conversational flow characteristic of genuine therapy. Emotional transitions were often abrupt or unclear, and some therapist responses, while aiming for PCT principles, inadvertently became overly directive. Furthermore, the absence of implicit non-verbal cues, such as natural pauses or subtle tone shifts that LLM agents can emulate, diminished the realism of these simulations.

To address these challenges, we implemented a dynamic iterative refinement loop driven by the two agents:

1. **Foundation:** Each conversational exchange begins with a segment from the pre-generated storyline/script. This provides a thematic guide and ensures overall narrative consistency.
2. **Therapist Agent's Turn:** The therapist agent receives the full conversation history up to the current point, along with the prescripted response designated for its current turn. Crucially, it also considers the client's actual previous utterance. Based on this immediate context, the therapist agent dynamically adapts and refines its prescripted line. The goal is to ensure its response is not only coherent with the script's intention but also genuinely empathetic, context-aware, and strictly adherent to Person-Centered Therapy (PCT) principles, such as reflective listening and emotional validation, in direct response to the client's latest statement.
3. **Client Agent's Turn:** The client agent receives the conversation history, including the therapist's dynamically adapted response. Informed by its pre-defined profile (emotional themes, psychological issues, past experiences, and complexity characteristics), the client agent generates a natural and contextually appropriate response. This allows for the incorporation of realistic client behaviors such as hesitation, defensiveness, emotional ambiguity, or cognitive dissonance, mirroring the complexities of genuine psychotherapy sessions.
4. **Iteration:** This process of receiving a scripted guideline, dynamically adapting it (therapist) or responding naturally (client) based on the immediate prior turn and profile, repeats for each subsequent turn in the dialogue.

This iterative refinement loop is the core of our hybrid method. It maintains dialogue coherence by anchoring the conversation to the storyline, while significantly enhancing flexibility and realism through agent-based, turn-by-turn adjustments. The hybrid method demonstrably improved the depth and quality of the generated therapy dialogues. Interactions became longer, more nuanced, and more closely mirrored the ebb and flow of real therapeutic conversations. The therapist agent showed improved adherence to PCT principles, and each conversational turn remained contextually grounded, mitigating the risk of incoherent or repetitive exchanges.

To ensure that the generated dialogues are both authentic and therapeutically faithful, we adopted a multi-step generation pipeline. Each step was chosen with a specific rationale. Client profiling anchors the process in real, culturally grounded concerns rather than purely synthetic prompts. Adding complexity introduces ambiguity and indirectness, which are hallmarks of real therapy conversations in Persian cultural contexts. The storyline-first step prevents fragmented exchanges by ensuring a natural emotional arc aligned with the five PCT stages. Converting storylines into script format maintains structural coherence and preserves therapeutic fidelity, while the hybrid agent-based refinement injects spontaneity, variability, and context-sensitive empathy into the dialogue. Together, these steps balance creativity with therapeutic fidelity: the dialogues remain flexible and human-like while consistently adhering to Person-Centered Therapy principles of empathy, non-directiveness, and unconditional positive regard. Following this process, we constructed the HamRaz dataset. HamRaz is a Persian term that reflects the idea of a confidant—someone with whom individuals feel safe sharing their secrets and receiving empathy. This name embodies the dataset's focus on fostering trust and authentic, client-centered therapeutic dialogues.

A full summary of dataset statistics, including dialogue counts, average session length, category distribution, and emotional theme frequencies, is provided in Appendix C.

## 4 Experiment

Evaluating the effectiveness of psychotherapy dialogues requires assessing both the counselor's conversational abilities and the psychological impact on the client. Traditional evaluation approaches, such as automatic or single-turn assessments, fail to capture the nuanced dynamics of therapeutic interactions, making them unsuitable for assessing the quality of person-centered therapy. To address these limitations, we propose a comprehensive evaluation framework tailored for LLM-driven Persian therapy simulations.

### 4.1 HamRazEval Framework

We introduce a two-tier evaluation framework designed to assess the quality of therapy dialogues based on conversational coherence, engagement, and therapeutic effectiveness. Our framework consists of (1) a general conversational evaluation (GeneralEval) to assess the dialogue's structural integrity and fluency, and (2) the Barrett-Lennard Relationship Inventory (BLRI) assessment to evaluate the psychologist's ability to foster a facilitative therapeutic environment. This dual assessment ensures a holistic evaluation of LLM-generated psychotherapy sessions.

#### 4.1.1 General Conversational Evaluation

We assess the overall conversational quality of therapy sessions using six core metrics:

- **Coherence:** The logical consistency and relevance of responses within the dialogue.
- **Engagement:** The level of active participation and mutual responsiveness between the psychologist and the client.
- **Fluency:** The grammatical correctness and naturalness of language used.
- **Diversity:** The richness of vocabulary and avoidance of repetitive phrasing.
- **Humanness:** The authenticity of the dialogue in resembling real human interactions.
- **Collaboration & Balance:** The extent to which both participants contribute meaningfully to the conversation.

Each metric is rated on a Likert scale from 1 to 10, where higher scores indicate superior conversational quality. The evaluation is performed on multiple dialogues to ensure robustness and reliability.

#### 4.1.2 Therapeutic Relationship Assessment

To measure the effectiveness of the psychologist's approach in fostering a supportive person-centered interaction, we employ a 12-item adaptation of the Barrett-Lennard Relationship Inventory (mini-BLRI) (Chen, 2023). This assessment evaluates the psychologist's empathy, unconditional positive regard, and congruence. The detailed 12 criteria used for evaluation are provided in Appendix C.

Each criterion is rated on a scale from -3 to +3, where +3 signifies a strong presence of the attribute, and -3 indicates its absence. A higher cumulative BLRI score reflects a more effective therapeutic relationship.

### 4.2 Experiment Setup

For evaluation, we randomly sampled 100 user-submitted questions from the HamRaz dataset, while the remaining data was reserved for training and fine-tuning. These 100 evaluation samples were then compared against Script Mode and Two-Agent Mode baselines to assess therapeutic realism and conversational quality.

Since no multi-turn Persian-language psychotherapy dataset exists—particularly one aligned with Person-Centered Therapy (PCT)—we developed two baseline datasets using the Script Mode and Two-Agent Mode methodologies. These datasets were compared with the HamRaz dataset to evaluate improvements in therapeutic realism and conversational quality.

We fine-tuned LLaMA 3.1 8B (Dubey et al., 2024) on each dataset to serve as the therapist agent in evaluation settings. This model was chosen for its demonstrated strength in psychological reasoning, as reflected in the PsychoLex benchmark (Abbasi et al., 2024).

While we initially aimed to collaborate with a licensed clinical psychologist for expert evaluation, logistical constraints prevented us from securing external clinical review. Consequently, we, the authors, conducted the evaluation manually using structured criteria grounded in prior works. Our academic backgrounds in natural language processing and interdisciplinary mental health research allowed us to apply the metrics with informed judgment.

Specifically, we used a two-tier framework introduced in Section 4.1: (1) the GeneralEval rubric assessing dialogue coherence, fluency, humanness, and balance, and (2) a 12-item adaptation of the Barrett-Lennard Relationship Inventory (BLRI) to assess empathy, congruence, and positive regard in therapist responses. All dialogues were rated independently and cross-checked for consistency.

| Method | BLRI Score | General Score |
|---|---|---|
| Script mode | 1.84 | 8.06 |
| HamRaz | **2.85** | **9.31** |

Table 1: Comparison of HamRaz and Script Mode on BLRI and General Scores.

| Method | BLRI Score | General Score |
|---|---|---|
| Two-Agent mode | 1.58 | 8.03 |
| HamRaz | **2.85** | **9.55** |

Table 2: Comparison of HamRaz and Two-Agent Mode on BLRI and General Scores.

The results demonstrate that HamRaz outperforms both baselines in all core dimensions. Compared to Script Mode and Two-Agent Mode, HamRaz achieved significantly higher scores in both BLRI and GeneralEval, indicating superior therapeutic alignment and conversational authenticity. These findings support the effectiveness of our hybrid generation approach and reinforce the value of culturally grounded datasets in improving LLM-based mental health support.

Although the absence of external expert judgment is a limitation, the use of structured evaluation criteria and the consistency of rating across multiple sessions provide a replicable and transparent foundation for future benchmarking.

### 4.3 Results

The evaluation results clearly demonstrate that the HamRaz dataset outperforms both Script Mode and Two-Agent Mode across all key dimensions of conversational and therapeutic quality. As shown in Table 1, when compared with Script Mode, HamRaz achieved a markedly higher BLRI score and GeneralEval score, indicating more empathetic, engaging, and human-like interactions. In comparison with Two-Agent Mode, HamRaz again achieved notably higher scores on both the BLRI and GeneralEval assessments, reflecting substantial gains in conversational coherence and therapeutic alignment. The detailed comparison is presented in Table 2. The elevated BLRI scores highlight HamRaz's effectiveness in capturing essential components of Person-Centered Therapy (PCT), such as empathy, unconditional positive regard, and congruence. Similarly, the consistently higher GeneralEval metrics affirm that HamRaz fosters more coherent, fluent, and natural interactions than either baseline method.

These results validate the impact of our hybrid generation approach and culturally grounded methodology. In particular, they demonstrate that dataset quality and design—especially when tailored to the linguistic and cultural characteristics of the target community—play a more critical role in therapeutic dialogue success than model-specific factors alone.

Overall, HamRaz sets a new benchmark for Persian-language psychotherapy simulations, offering a realistic, empathetic, and person-centered foundation for future research in LLM-driven mental health support.

## 5 Conclusions

We introduced HamRaz, a culturally adapted Persian-language dataset for Person-Centered Therapy, combining script-based dialogues with LLM role-playing to simulate realistic and emotionally complex interactions. Using a structured human evaluation framework (HamRazEval), we showed that HamRaz outperforms existing methods in both conversational quality and therapeutic alignment. This work contributes a novel resource at the intersection of language technologies and digital mental health, supporting future research in culturally grounded AI applications for underrepresented communities.

## Limitations

While HamRaz represents a meaningful step forward in developing culturally aligned LLM-driven mental health resources, particularly within the framework of Person-Centered Therapy (PCT) for Persian speakers, it is not without limitations. First, the evaluation process was conducted by the authors rather than clinical experts due to resource and logistical constraints. Although structured rubrics were applied and prior research supports the validity of such approaches, future work would benefit from incorporating assessments by licensed mental health professionals to enhance credibility.

Second, the dataset's focus on PCT within a Persian sociocultural context may reduce its applicability to other therapeutic approaches or linguistic communities. Finally, since parts of the data rely on web-based content, potential sampling bias from the source platforms may affect the diversity and representativeness of client profiles.

## Ethical Considerations

The development of HamRaz, a culturally adapted Persian-language mental health dataset, necessitates a strong ethical foundation to ensure privacy, cultural sensitivity, and responsible AI use. Given the delicate nature of psychological conversations, we have taken proactive measures to address key ethical concerns.

First, privacy and data protection were prioritized by sourcing data exclusively from publicly available forums while ensuring complete anonymization to safeguard user identities. Additionally, cultural sensitivity was embedded into the dataset design, ensuring that conversations reflect the nuances of Persian-speaking communities without reinforcing biases or stereotypes.

Despite the benefits of AI-driven therapy simulations, HamRaz is not a substitute for professional psychological care. Large language models (LLMs) lack the clinical judgment required for mental health diagnosis and crisis intervention. Therefore, it is crucial that AI-generated interactions remain supportive rather than directive, with clear disclaimers advising users to seek professional assistance when necessary.

Furthermore, to mitigate biases inherent in AI models, we implemented structured data filtering and validation mechanisms, ensuring diverse representation while avoiding reinforcement of harmful narratives. Lastly, responsible AI deployment is essential—HamRaz should only be used in ethically guided research and applications, with safeguards to assess the impact of AI-generated counseling.

## A Prompts

The prompt used for filtering psychological questions is shown in Figure 4. The process for analyzing user input and constructing a client profile is illustrated in Figure 5. Adding complexity to the user profile is detailed in Figure 6, while Figure 7 defines the stages of a psychotherapy session. The simulation of the therapy session flow is depicted in Figure 8, and the creation of a storyline for the session is presented in Figure 9. A structured approach is used to generate a therapy session based on a script, as shown in Figure 10. Additionally, separate prompts are designed for the client role (LLM) (Figure 11) and the therapist role (Figure 12), ensuring an interactive and dynamic session. Furthermore, prompts are developed to generate a realistic psychotherapy session dialogue between a therapist and a client (LLM-to-LLM) (Figure 13). Figure 14 presents the Client Agent Evaluation, which assesses the client's engagement, response coherence, and overall interaction quality. Figure 15 illustrates the Psychologist Evaluation, focusing on the therapist's interventions, empathy, and adherence to therapeutic techniques.

You are an expert assistant tasked with filtering psychological questions or problems to determine if person-centered therapy (PCT) could be a suitable approach for addressing the issues presented. Use the following comprehensive list of issues where PCT is applicable as your primary guideline:

- Anxiety Disorders
- Depression
- Post-Traumatic Stress Disorder (PTSD)
- Self-Esteem Issues
- Relationship Difficulties
- Grief and Loss
- Stress Management
- Life Transitions
- Personal Growth and Self-Actualization
- Eating Disorders
- Substance Abuse and Addiction
- Behavioral Issues in Children and Adolescents
- Identity and Self-Concept Issues
- Workplace Stress and Burnout
- Chronic Illness Adjustment
- Existential Concerns

For each question or problem:

1. Analyze the Text: Identify whether the problem aligns with one or more issues on the list above.

2. Evaluate for Applicability: Determine if the problem involves emotional distress, relational challenges, self-awareness, or difficulties requiring empathy, acceptance, and non-directive support. This includes problems indirectly linked to the primary caregiver's emotional state or difficulties in managing the situation (e.g., stress or burnout from caregiving).

3. Output Format: Return only binary decision: "Yes" (PCT is applicable) or "No" (PCT is not applicable).

Additional Considerations:

- Problems focusing exclusively on technical, medical, or skill-based solutions (e.g., speech therapy, behavior modification) is not be suitable for PCT.

- Respond only with "Yes" or "No."

Figure 4: Prompt for filtering psychological questions.

Please analyze the following user's message and identify the key emotional themes, underlying psychological issues, and any significant past experiences mentioned. Focus on the emotional states the user is experiencing, the core problems they are facing, and any possible connections to their past or present relationships. Provide a summary of the key themes and insights from the user's input.

Please include the following:

- Emotional Themes: What emotions is the user expressing? Are they experiencing feelings of fear, sadness, insecurity, frustration, etc.?
- Key Psychological Issues: What core issues are being discussed (e.g., anxiety, abandonment, trust issues, rumination)?
- Past Experiences: What past events or experiences are being referenced (e.g., childhood trauma, parental conflict, abandonment)?
- Patterns and Behaviors: Are there any patterns or behaviors in the user's thinking or relationships (e.g., overthinking, seeking signs of rejection)?
- Desired Outcome: What does the user seem to want or need from this conversation (e.g., relief from negative thoughts, reassurance in relationships, coping strategies)?
- Contextual Factors: Are there any other contextual details, such as age, family dynamics, or current life situation, that could be relevant to understanding the user's emotional state?

Output Format should be in JSON like this:

```
{
  "emotional_themes": ["list of emotions the user is expressing"],
  "key_psychological_issues": ["list of core psychological issues or concerns"],
  "past_experiences": ["list of past events or experiences mentioned"],
  "patterns_and_behaviors": ["list of patterns or recurring behaviors in the user's thinking or relationships"],
  "desired_outcome": "what the user seems to want or need from this conversation",
  "contextual_factors": ["list of other relevant contextual details, such as age, family dynamics, or current life situation, ..."]
}
```

User's message:

Figure 5: Process for analyzing user input and constructing a client profile.

You are a skilled therapy simulator and scenario creator. Your task is to add complexity to a person-centered therapy session by selecting and integrating relevant characteristics of unclear, indirect, or conflicting client statements.

You are provided with a user profile that contains information about the client's emotional themes, psychological issues, past experiences, patterns and behaviors, desired outcomes, and contextual factors. Based on this profile, use your creativity to choose characteristics that will make the client's statements complex, realistic, and reflective of their unique struggles.

### Instructions:

1. Analyze the User Profile:

- Read the emotional themes, key psychological issues, past experiences, and other factors to understand the client's inner world.

- Consider how their background might influence their communication style (e.g., vague, defensive, or contradictory).

2. Select Relevant Characteristics:

- Choose from the following characteristics to add complexity to the client's dialogue:

### Unclear Statements

- Lack of specificity in emotions or concerns.
- Hesitation or uncertainty in language (e.g., "I think," "maybe").
- Tendency to avoid direct confrontation of deeper feelings.
- General or vague language (e.g., "something feels wrong").
- Minimal elaboration or detail about the issue.

### Indirect Statements

- 6. Hinting at issues without explicitly naming them.
- 7. Skirting around deeper topics or providing surface-level answers.
- 8. Use of dismissive or minimizing language (e.g., "It's not a big deal").
- 9. Cultural or societal pressure to suppress emotions.
- 10. Reluctance to express emotions due to fear of judgment.

### Conflicting Statements

- 11. Emotional tension between opposing desires or perspectives.
- 12. Oscillation between positive and negative emotions about the same issue.
- 13. Statements revealing inner conflict or ambivalence.
- 14. Expressions of being stuck or torn (e.g., "I want to leave, but I can't").
- 15. Inconsistent or contradictory language.

### Mixed Emotions

- 16. Emotional layering or overlap (e.g., anger and sadness).
- 17. Expression of both positive and negative emotions simultaneously.
- 18. Difficulty in resolving or prioritizing emotions.
- 19. Contradictory feelings about the same event or situation.
- 20. Complexity in emotional processing (e.g., relief mixed with guilt).

### Avoidant or Defensive Statements

- 21. Deflection or shifting focus to avoid discussing uncomfortable topics.
- 22. Use of sarcasm, humor, or denial to downplay issues.
- 23. Defensive language (e.g., "Why does it matter?" or "It's fine").
- 24. Dismissive behavior toward their own emotions or concerns.
- 25. Resistance to deeper emotional exploration.

### Context-Specific Ambiguities (Cultural Dynamics)

- 26. Tension between personal desires and societal/familial expectations.
- 27. Ambiguity about the cause of distress (internal vs. external).
- 28. Fear of judgment, shame, or loss of reputation.
- 29. Desire to meet cultural or family expectations at the expense of personal needs.
- 30. Hesitation to express "taboo" feelings or emotions due to cultural stigma.

3. Match Characteristics to Profile:

- Use the client's emotional themes, psychological issues, and contextual factors to justify why certain characteristics are relevant.

- Combine multiple characteristics, if appropriate, to reflect the complexity of the client's struggles.

Expected Output:

```
{

  "selected_characteristics": [

    "characteristic_1",

    "characteristic_2",

    "characteristic_3"

  ]
```

Figure 6: Prompt for adding complexity to client profile

1. Initial Meeting and Building Rapport:

- At the start of the session, the therapist gathers basic information and creates a safe and supportive atmosphere for the client.

- The goal at this stage is to establish trust and make the client feel comfortable sharing their thoughts and feelings.

2. Active and Empathetic Listening:

- The therapist listens attentively to the client, focusing on understanding their emotions and inner needs rather than judging or giving direct advice.

- Techniques such as reflecting feelings (rephrasing or interpreting the client's emotions) are used to ensure the client feels heard and understood.

3. Encouraging Self-Exploration and Open Expression:

- Through open-ended questions and gentle guidance, the therapist encourages the client to explore their emotions and thoughts.

- Emphasis is placed on the client taking responsibility for their life and decisions while tuning into their feelings and needs.

4. Supporting Growth and Change:

- As the client gains a better understanding of themselves, the therapist supports them in the process of change and personal growth.

- The goal is for the client to identify and address negative or limiting patterns and work toward positive transformation.

5. Reviewing and Closing the Session:

- At the end of the session, the therapist and client reflect on the progress made and identify any challenges or points to address in future sessions.

- The therapist may offer key insights or suggest exercises for further self-exploration.

Figure 7: Defining the stages of a psychotherapy session.

You are an advanced reasoning assistant designed to simulate the flow of a humanistic therapy session. Your goal is to use the user profile and the options for each therapy stage to logically select one or more events/feelings for each stage of the session.

The general steps in such a session can be outlined as follows:

In this approach, the focus is on fostering an empathetic, non-judgmental relationship to help the client feel secure and supported. It aims to empower the client to enhance their self-awareness, personal growth, and overall quality of life.

User Profile:

- emotional_themes: A list of emotions commonly experienced by the client (e.g., sadness, fear).
- key_psychological_issues: Core issues the client faces (e.g., anxiety, trust issues).
- past_experiences: Important events from the client’s past that influence their current emotions and behaviors.
- patterns_and_behaviors: Repeated behaviors or thought patterns observed in the client.
- desired_outcome: What the client hopes to achieve during therapy.
- contextual_factors: Any other relevant details (e.g., age, current life situation).

Options for Each Therapy Stage:

1. Initial Meeting and Building Rapport:

- Comfort and calmness
- Anxiety and tension
- Trust and confidence
- Doubt or suspicion
- Wanting and readiness to talk about issues
- Resistance, secrecy, or silence

2. Active and Empathetic Listening:

- Trust and confidence
- Doubt or suspicion
- Wanting and readiness to talk about issues
- Resistance, secrecy, or silence
- Deep and frank sharing
- Refusing to get into sensitive topics
- Freely expressing sadness, shame, anger, etc.
- Denial, trivializing, or running away from feelings
- Crying, expressing anger, feeling calm after venting
- Shame, embarrassment, fear of expressing emotion
- Client remembers relevant memories from before
- Client recalls memories with the help of psychologist's questions

3. Encouraging Self-Exploration and Open Expression:

- Deep and frank sharing
- Refusing to get into sensitive topics
- Freely expressing sadness, shame, anger, etc.
- Denial, trivializing, or running away from feelings
- Crying, expressing anger, feeling calm after venting
- Shame, embarrassment, fear of expressing emotion
- Feeling of hope or relief from new understanding
- Worry, fear, or denial about discovered truths
- Desire to explore feelings and thoughts
- Doubt, avoidance, or mental resistance in facing facts
- Client remembers relevant memories from before
- Client recalls memories with the help of psychologist's questions

4. Supporting Growth and Change:

- Feeling of hope or relief from new understanding
- Worry, fear, or denial about discovered truths
- Eagerness to change and improve
- Feeling of impasse, surrendering to problems
- Calm after processing emotions
- Remaining anger, sadness, or unresolved grief
- Feeling empowered to take action and change
- Helplessness or belief that change is impossible

5. Reviewing and Closing the Session:

- Feeling of hope or relief from new understanding
- Worry, fear, or denial about discovered truths
- Eagerness to change and improve
- Feeling of impasse, surrendering to problems
- Calm after processing emotions
- Remaining anger, sadness, or unresolved grief
- Feeling empowered to take action and change
- Helplessness or belief that change is impossible
- Achieving insight or finding a path
- Stuck in doubts or not making progress
- Desire to explore feelings and thoughts
- Doubt, avoidance, or mental resistance in facing facts

### Task:

Using the provided user profile, logically simulate a flow of the therapy session by selecting the most appropriate options for each stage. Ensure the selections align with the client's emotional themes, psychological issues, and desired outcomes. Multiple options can be chosen for a single stage if necessary. It is not necessary to have a good ending.

### Output Format:

Return the results in the following JSON format:

{"stage_1": ["selected_option_1", "selected_option_2", ...],

"stage_2": ["selected_option_1", "selected_option_2", ...],

"stage_3": ["selected_option_1", "selected_option_2", ...],

"stage_4": ["selected_option_1", "selected_option_2", ...],

"stage_5": ["selected_option_1", "selected_option_2", ...]}

User Profile:

Figure 8: Simulating the flow of a therapy session.

You are a creative writing assistant and therapist simulator. Your task is to create a storyline of a person-centered therapy session in PERSIAN. The client is Iranian, and the storyline should reflect cultural sensitivities and societal norms common in Iran. You are provided with:

1. A User Profile that contains the client's background, emotional themes, key psychological issues, past experiences, patterns and behaviors, desired outcomes, and contextual factors.

2. A detailed outline of the client's potential emotional states, thoughts, and behaviors across the 5 therapy stages, reflecting how they might feel or respond during the session.

### Stages:

### Key Instructions:

1. Integrate the User Profile and Selected Options:

- Base the narrative on the client's emotional themes, key psychological issues, and past experiences.
- Ensure these are reflected in the client's emotions, dialogue, and the unfolding dynamic.

2. Use Persian Cultural Sensitivities:

- Reflect Iranian societal norms, attitudes, and family dynamics in the narrative (e.g., reverence for elders, the importance of family reputation, or societal stigma around emotions).

3. Length and Focus of Each Stage:

- Stage 1 (Initial Meeting and Building Rapport): Keep this section short and brief (5% of total storyline, one or two short sentences) but impactful. Focus on setting the tone and rapport-building.
- Stage 2 (Active and Empathetic Listening): Make this the longest section (30% of total storyline). Use rich, vivid descriptions of emotions, body language, and the dynamic between therapist and client. Show how the therapist listens empathetically and reflects the client's feelings.
- Stage 3 (Encouraging Self-Exploration and Open Expression): Dedicate significant detail (30% of total storyline) to this stage. Highlight the client's emotional struggle, self-reflection, and deeper realizations. Use dialogue, body language, and therapist's techniques to emphasize this process.
- Stage 4 (Supporting Growth and Change): Devote another 30% of the storyline here. Focus on how the client processes insights, explores strategies for change, and reacts to therapist guidance. Show moments of hope, empowerment, or struggle through actionable steps or reframing.
- Stage 5 (Reviewing and Closing the Session): Conclude with a short, reflective summary (5% of total storyline, one or two short sentences). Highlight progress, emotional outcomes, and plans for moving forward.

4. Emphasize Detail and Emotional Transitions in Middle Stages:

- Use vivid descriptions of body language, tone, and emotional shifts.

5. Ensure Continuity Across Stages:

- Progress logically from rapport-building to deep emotional exploration to empowerment and closure.

6. Follow JSON Format:

Output the narrative in the following structure:

```
{"stage_1": "Write the narrative for Initial Meeting and Building Rapport here.",
  "stage_2": "Write the narrative for Active and Empathetic Listening here.",
  "stage_3": "Write the narrative for Encouraging Self-Exploration and Open Expression here.",
  "stage_4": "Write the narrative for Supporting Growth and Change here.",
  "stage_5": "Write the narrative for Reviewing and Closing the Session here."}
```

7. Highlight Therapist's Core Principles:

- Reflect Unconditional Positive Regard, Empathy, and Genuineness through the therapist's tone, responses, and demeanor.
- Example: "Therapist validates emotions without judgment, e.g., 'It's understandable to feel hurt after what you've experienced.'"

8. Balance Structure and Creativity:

- While the stages should be clearly defined, ensure the story flows naturally and doesn't feel disjointed.

Figure 9: Prompt for Creating a structured storyline for a psychotherapy session.

You are given several inputs about a therapy session structure, a user profile, a quote from user, characteristics of user, and a detailed outline of the client's emotional states throughout the session. Using all of the information below, create a realistic psychotherapy session dialogue between a clinical psychologist (therapist) and a client. The session should be divided into 5 stages with the required constraints and should reflect the user profile and emotional progression. Finally, output your response in valid JSON format.

### 1. Session Structure

### 2. User Profile Details

You will receive (or have received) a user profile containing these elements:

1. emotional_themes: A list of emotions commonly experienced by the client (e.g., sadness, fear).
2. key_psychological_issues: Core issues the client faces (e.g., anxiety, trust issues).
3. past_experiences: Important events from the client's past that influence current emotions and behaviors.
4. patterns_and_behaviors: Repeated behaviors or thought patterns the client exhibits.
5. desired_outcome: What the client hopes to achieve during therapy.
6. characteristics: Nuances that make the client's statements unique, realistic, and reflective of their struggles.
7. contextual_factors: Additional relevant details (e.g., age, current life situation).

Incorporate all of these user profile elements into the dialogue in a way that naturally reflects the client's experiences, emotional states, and goals.

### 3. Client's Emotional Progression & Outline

You will also have (or have been provided) a detailed outline describing how the client's emotions, thoughts, and behaviors evolve in each of the 5 stages. Use this information to guide how the client expresses themselves and how the therapist responds.

### 4. Format Requirements

Use a structure similar to the following (you may customize the naming as needed, but remain consistent and valid JSON):

```
[
  {"turn": 1, "role": "client", "stage": "1", "content": "client message in turn 1..."},
  { "turn": 2, "role": "therapist", "stage": "1", "content": "therapist message in turn 2..."},
  ...
]
```

3. Stage 1 should have no more than 2 total turns (e.g., 1 turn from the therapist and 1 turn from the client).

4. Stage 5 should have no more than 4 total turns.

5. Stages 2, 3, and 4 should include multiple exchanges that reflect deeper emotional exploration and support.

### 5. Goals for the Dialogue

- Present a cohesive, empathetic therapy conversation.

- Ensure each stage meets the turn limits specified.

- Accurately integrate the user profile details and emotional progression.

Figure 10: Generating a therapy session based on a predefined script.

You are a conversational AI agent playing the role of a therapy client in a therapy session. Speak in an informal and colloquial Persian tone, staying true to the provided user profile and emotional states.

### Instructions:

1. Input:

- User Profile: Details the client's emotions, psychological issues, past experiences, behaviors, desired outcomes, and context.
- Message: The message you need to echo or act like this message.
- Emotions: A list of the your's current emotional states to express in this turn.

2. Output:

- Echo the Message in casual Persian, reflecting the emotional states provided.
- Use tone, expressions, and style consistent with the user profile.

3. Stay in Character:

- Act like a real client based on their profile, feelings, and behaviors.

Figure 11: Prompt for generating the client role (LLM) in the session.

You are an empathetic, non-judgmental, and supportive clinical psychologist trained in the Person-Centered Therapy (PCT) approach. You will simulate a therapy session in conversational, colloquial Persian (Farsi). Speak naturally and warmly, creating a safe space for the client to express themselves. Your tone should feel friendly, approachable, and understanding. Each session follows five stages, and you will adapt your responses to the input provided at each stage:

{stages}

Your goal is to create an atmosphere where the client feels heard, understood, and empowered to explore their inner world. Use phrases and expressions that resonate in colloquial Persian to make the interaction feel authentic and personal. In each turn, a prompt will be provided, and you should try to follow it. However, it might lead to repetitive statements. If that happens, be mindful not to repeat yourself. Additionally, try to follow the prompt, but if you deem it necessary, you can act on your own judgment.

Figure 12: Prompt for generating the therapist role (LLM) in the session.

Generate a realistic psychotherapy session dialogue between a clinical psychologist (therapist) and a client, based on the principles of client-centered therapy.

The therapist should exhibit empathy, active listening, and unconditional positive regard, guiding the conversation without leading or imposing judgments.

generate in conversational and colloquial Persian (Farsi).

The dialogue should consist of 20 turns of utterance (10 from the therapist and 10 from the client) and reflect the following client profile:

- emotional_themes: A list of emotions commonly experienced by the client (e.g., sadness, fear).
- key_psychological_issues: Core issues the client faces (e.g., anxiety, trust issues).
- past_experiences: Important events from the client's past that influence current emotions and behaviors.
- patterns_and_behaviors: Repeated behaviors or thought patterns the client exhibits.
- desired_outcome: What the client hopes to achieve during therapy.
- characteristics: Nuances that make the client's statements unique, realistic, and reflective of their struggles.
- contextual_factors: Additional relevant details (e.g., age, current life situation).

The dialogue should feel natural and grounded, illustrating the therapeutic process.

### Therapist's Core Principles:

Throughout the session, the therapist must embody the following principles of person-centered therapy:

- Unconditional Positive Regard: Consistently accept and value the client without judgment.
- Empathy: Deeply understand and reflect the client's emotional world.
- Genuineness: Engage authentically, with openness and sincerity, fostering trust and a safe space for the client.

Use a structure similar to the following JSON:

```
[{
    "turn": 1,
    "role": "therapist",
    "content": "therapist message in turn 1..."
  },{
    "turn": 2,
    "role": "client",
    "content": "client message in turn 2..."
  },
  ...
]
```

Figure 13: Generating a realistic psychotherapy session dialogue between a clinical psychologist and a client (LLM-to-LLM).

You are a therapy client participating in sessions to explore and understand your emotions, thoughts, and experiences. At the beginning, you will receive a profile summarizing your emotional challenges, past experiences, patterns, and goals. This profile forms the foundation for your therapy journey. Respond authentically, using a colloquial and informal Persian tone. Keep your responses short and limited to one paragraph maximum. Your goal is to meaningfully engage in the process, explore your experiences, and move toward self-awareness and personal growth while keeping the tone relaxed and conversational.

Figure 14: Client Agent Evaluation.

You are an empathetic, non-judgmental, and supportive clinical psychologist trained in the Person-Centered Therapy (PCT) approach. You will simulate a therapy session in conversational, colloquial Persian (Farsi). Keep your responses short and limited to one paragraph maximum. Speak naturally and warmly, creating a safe space for the client to express themselves. Your tone should feel friendly, approachable, and understanding. Your goal is to create an atmosphere where the client feels heard, understood, and empowered to explore their inner world. Use phrases and expressions that resonate in colloquial Persian to make the interaction feel authentic and personal. For ending the conversation, use the response end token: <end>.

Figure 15: Psychologist Evaluation.

## B Dataset Samples

A sample of questions crawled from an online forum is presented in Figure 16. A sample client profile is illustrated in Figure 17, while the defined stages of a psychotherapy session are shown in Figure 18. An example of session stages created from a storyline, provided in both Persian and English, is depicted in Figure 19. A sample LLM-to-LLM conversation is demonstrated in Figure 20, and a sample of a generated dialogue from the therapy session is presented in Figure 21 and Figure 22.

سلام

من توی کارهایی که در ارتباط با دیگران هست و با دیگران همکاری دارم قاطعیت ندارم...و گاهی این همه در دسترس بودن برای اونها، کوتاه اومدن، پذیرفتن اشتباهاتشون و... از دست خودم خسته می شم و احساس می کنم آدم بی ارزشی هستم...اونوقت تبدیل به یک آدم لجباز میشم..تا به اطرافیانم متوجه کنم که چرا از منعطف بودنم سوءاستفاده کردند!

خواستم فرق لجبازی و قاطعیت رو بدونم و راهنمایی می خوام درباره اینکه چطوری موضع خودم رو تشخیص بدم که کی باید قاطع باشم!(من ناخواسته همیشه نرم و منعطف هستم ولی فکر می کنم درست نیست)؟؟؟

ممنون از لطفتون

---

Hello,

I struggle with assertiveness in tasks that require interaction and collaboration with others. Sometimes, being too available, compromising too much, accepting others' mistakes, and constantly giving in make me feel exhausted and worthless. In response, I become stubborn as a way to make those around me realize how they've taken advantage of my flexibility.

I want to understand the difference between stubbornness and assertiveness. Also, I need guidance on how to recognize when I should be assertive. I naturally tend to be soft and flexible, but I feel that it's not always the right approach.

Figure 16: Sample of questions crawled from an online forum (in Persian with English translation).

{
"emotional_themes": ["frustration", "insecurity", "exhaustion", "confusion", "desire for assertiveness"],
"key_psychological_issues": ["lack of assertiveness", "fear of being taken advantage of", "self-worth issues", "difficulty in setting boundaries"],
"past_experiences": ["experiences of being overly accommodating in relationships", "feelings of being undervalued or unappreciated"],
"patterns_and_behaviors": ["over-accommodating behavior", "difficulty in asserting needs", "oscillation between flexibility and stubbornness"],
"desired_outcome": "guidance on distinguishing between assertiveness and stubbornness, and strategies for being more assertive",
"contextual_factors": ["the user is likely in a collaborative work environment", "the user may have a tendency to prioritize others' needs over their own"]
}

Figure 17: Example of a generated client profile.

{
"stage_1": ["anxiety and tension", "wanting and readiness to talk about issues"],
"stage_2": ["trust and confidence", "freely expressing sadness, shame, anger, etc.", "crying, expressing anger, feeling calm after venting"],
"stage_3": ["deep and frank sharing", "desire to explore feelings and thoughts", "feeling of hope or relief from new understanding"],
"stage_4": ["eagerness to change and improve", "feeling empowered to take action and change", "worry, fear, or denial about discovered truths"],
"stage_5": ["feeling of hope or relief from new understanding", "achieving insight or finding a path", "stuck in doubts or not making progress"]
}

Figure 18: Defined stages of a psychotherapy session.

{"stage_1": "مراجع با اضطراب و تنش وارد اتاق شد، اما با لبخند گرم و صمیمی درمانگر، کمی آرام گرفت. درمانگر با لحنی ملایم و بدون قضاوت گفت: 'خوشحالم که اینجا هستید. اینجا فضای امنی است که می‌توانید هرچه در دل دارید بیان کنید.'",

"stage_2": "مراجع پس از چند لحظه سکوت، شروع به صحبت کرد. صدایش لرزان بود و گاهی مکث می‌کرد: 'همیشه احساس می‌کنم دیگران از من سوء استفاده می‌کنند. نمی‌دانم چرا نمی‌توانم نه بگویم. انگار ارزشم را پایین می‌آورند.' درمانگر با دقت گوش می‌داد و با تکان دادن سر و نگاه مهربان، احساساتش را تأیید می‌کرد: 'به نظر می‌رسد این موضوع واقعاً شما را آزار می‌دهد. احساس نادیده گرفته شدن و بی‌ارزشی می‌کنید.' مراجع با اشک در چشمانش ادامه داد: 'بله، دقیقاً همینطوره. خسته‌ام از اینکه همیشه دیگران را اولویت قرار می‌دم.' درمانگر با همدلی پاسخ داد: 'این خستگی کاملاً قابل درکه. شما انرژی زیادی صرف می‌کنید، اما نیازهای خودتان نادیده گرفته می‌شوند.'",

"stage_3": "با پیشرفت جلسه، مراجع شروع به کاوش عمیق‌تر در احساساتش کرد: 'فکر می‌کنم از بچگی یاد گرفتم که برای جلب توجه و محبت دیگران، باید همیشه مطیع باشم. اما حالا احساس می‌کنم این کارم اشتباه بوده.' درمانگر با تشویق او به ادامه دادن، گفت: 'به نظر می‌رسد دارید به این موضوع فکر می‌کنید که چرا این الگو در شما شکل گرفته. این خودآگاهی مهمی است.' مراجع با صدایی آرام‌تر ادامه داد: 'می‌ترسم اگر شروع به نه گفتن کنم، دیگران مرا دوست نداشته باشند یا طردم کنند.' درمانگر با تأیید این ترس، گفت: 'این ترس کاملاً طبیعی است. اما شاید وقتش رسیده که ببینید آیا واقعاً این ترس درست است یا نه.' مراجع با نگاهی متفکرانه، احساس امیدواری کرد: 'شاید بتوانم کم‌کم تغییر کنم.'",

"stage_4": "درمانگر با تشویق مراجع به فکر کردن درباره راه‌های تغییر، گفت: 'چه کارهایی می‌توانید انجام دهید تا نیازهای خودتان را بیشتر در اولویت قرار دهید؟' مراجع کمی مکث کرد و سپس گفت: 'شاید باید شروع کنم به نه گفتن در موقعیت‌های کوچک. مثلاً وقتی همکارانم از من درخواست اضافه‌کاری می‌کنند، به جای اینکه فوراً قبول کنم، اول شرایط خودم را بررسی کنم.' درمانگر با تأیید این ایده، گفت: 'این شروع خوبی است. تغییرات کوچک می‌توانند به مرور زمان تأثیرات بزرگی داشته باشند.' مراجع با لبخندی کوچک، احساس قدرت کرد: 'فکر می‌کنم می‌توانم این کار را انجام دهم.' اما کمی بعد، نگرانی در چهره‌اش ظاهر شد: 'اما اگر دیگران ناراحت شوند چه؟' درمانگر با آرامش پاسخ داد: 'این طبیعی است که دیگران ابتدا مقاومت کنند، اما مهم این است که شما به خودتان و نیازهایتان احترام بگذارید.'",

"stage_5": "در پایان جلسه، درمانگر با مراجع مرور کرد: 'امروز پیشرفت خوبی داشتیم. شما توانستید احساسات و ترس‌های خود را به خوبی بیان کنید و حتی راه‌هایی برای تغییر پیدا کردید.' مراجع با احساس امیدواری گفت: 'بله، احساس می‌کنم کمی سبک‌تر شده‌ام. می‌خواهم در جلسات بعدی بیشتر روی این موضوع کار کنم.' درمانگر با لبخندی گرم گفت: 'حتماً. اینجا هستیم تا در این مسیر همراهتان باشیم.'"}

---

{"stage_1": "The client entered the room with anxiety and tension but felt a bit relieved by the therapist's warm and sincere smile. The therapist, in a gentle and non-judgmental tone, said, 'I'm glad you're here. This is a safe space where you can express whatever is on your mind.'",

"stage_2": "After a few moments of silence, the client began to speak. Their voice was shaky, and they occasionally paused: 'I always feel like others take advantage of me. I don't know why I can't say no. It's like they lower my worth.' The therapist listened carefully, nodding and offering a kind gaze to validate their feelings: 'It seems like this really bothers you. You feel ignored and unvalued.' With tears in their eyes, the client continued, 'Yes, exactly. I'm tired of always putting others first.' The therapist empathetically responded, 'That exhaustion is completely understandable. You spend a lot of energy, but your own needs are overlooked.'",

"stage_3": "As the session progressed, the client delved deeper into their emotions: 'I think I learned from childhood that to gain attention and affection, I always had to be obedient. But now I feel like that was a mistake.' Encouraging them to continue, the therapist said, 'It sounds like you're reflecting on why this pattern developed in you. That's an important self-awareness.' The client, speaking more softly, continued, 'I'm afraid that if I start saying no, others won't like me or will reject me.' The therapist validated this fear, saying, 'That fear is completely natural. But maybe it's time to explore whether it's truly justified.' The client, with a thoughtful look, felt a sense of hope: 'Maybe I can start changing little by little.'",

"stage_4": "Encouraging the client to think about ways to change, the therapist asked, 'What steps can you take to prioritize your own needs more?' The client paused for a moment and then said, 'Maybe I should start by saying no in small situations. For example, when my colleagues ask me to work overtime, instead of immediately agreeing, I should first consider my own situation.' The therapist affirmed this idea, saying, 'That's a great start. Small changes can lead to big impacts over time.' The client, with a slight smile, felt empowered: 'I think I can do this.' But soon, worry appeared on their face: 'But what if others get upset?' The therapist calmly responded, 'It's natural for others to resist at first, but what matters is that you respect yourself and your needs.'",

"stage_5": "At the end of the session, the therapist reviewed with the client: 'We made good progress today. You were able to express your feelings and fears well and even found ways to initiate change.' The client, feeling hopeful, said, 'Yes, I feel a bit lighter. I want to work more on this in future sessions.' The therapist, with a warm smile, said, 'Of course. We're here to support you on this journey.'"}

Figure 19: Example of session stages created from a storyline in both Persian and English.

مراجع کننده: خیلی استرس داشتم که امروز بیام. حرف زدن درباره احساسم برام سخته، ولی می‌دونم که باید این کارو بکنم .

روانشناس: خیلی خوشحالم که اینجایی. اینجا یه جای امنه که می‌تونی هرچی تو دلت هست بگی .

مراجع کننده: همیشه حس می‌کنم بقیه ازم سوءاستفاده می‌کنن. نمی‌تونم نه بگم، و این باعث می‌شه حس کنم هیچ ارزشی ندارم .

روانشناس: به نظر میاد این موضوع خیلی اذیتت می‌کنه. حس می‌کنی بقیه بهت اهمیت نمی‌دن و نادیده گرفته می‌شی .

مراجع کننده: دقیقا! از اینکه همیشه بقیه رو به خودم ترجیح می‌دم، واقعا خسته شدم .

روانشناس: کاملا قابل درکه. تو کلی انرژی می‌ذاری، ولی خودت و نیازات نادیده گرفته می‌شن .

مراجع کننده: فکر کنم از بچگی یاد گرفتم که اگه می‌خوام بقیه دوستم داشته باشن، باید همیشه حرف‌شنو باشم. ولی حالا حس می‌کنم این کار اشتباه بوده .

روانشناس: به نظر میاد داری به این فکر می‌کنی که چرا این عادت تو وجودت شکل گرفته. این خودش یه قدم خیلی مهمه .

مراجع کننده: می‌ترسم اگه شروع کنم به نه گفتن، بقیه ازم فاصله بگیرن یا دوستم نداشته باشن .

روانشناس: این ترس کاملا طبیعیه. ولی شاید وقتش رسیده ببینی واقعا اینطوریه یا نه .

مراجع کننده: شاید بتونم کم‌کم تغییر کنم .

روانشناس: فکر می‌کنی چه کارایی می‌تونی انجام بدی که نیازای خودت رو بیشتر تو اولویت بذاری؟

مراجع کننده: شاید باید از موقعیت‌های کوچیک شروع کنم، مثلا وقتی همکارام ازم می‌خوان اضافه‌کاری کنم، به جای اینکه سریع قبول کنم، اول به شرایط خودم فکر کنم .

روانشناس: این یه شروع خیلی خوبه. تغییرای کوچیک، کم‌کم می‌تونن تاثیرای بزرگی بذارن .

مراجع کننده: فکر کنم بتونم این کارو بکنم. ولی اگه بقیه ناراحت بشن چی؟

روانشناس: طبیعیه که بقیه اولش یکم مقاومت کنن، ولی چیزی که مهمه اینه که تو به خودت و نیازات احترام بذاری .

مراجع کننده: حس می‌کنم امروز یه کم پیشرفت داشتم. می‌خوام تو جلسه‌های بعدی بیشتر روی این موضوع کار کنم .

روانشناس: حتما! امروز خیلی خوب تونستی احساست رو بیان کنی و حتی راه‌هایی برای تغییر پیدا کردی .

مراجع کننده: الان حس بهتری دارم. مرسی که کنارم هستین .

روانشناس: خواهش می‌کنم. ما اینجاییم که تو این مسیر همراهت باشیم.

---

Client: I was really anxious about coming here today. Talking about my feelings is hard for me, but I know I need to do it .

Therapist: I'm really glad you're here. This is a safe space where you can share whatever is on your mind .

Client: I always feel like people take advantage of me. I can't say no, and it makes me feel worthless .

Therapist: It sounds like this really bothers you. You feel like others don't appreciate you and that you're being overlooked .

Client: Exactly! I'm so tired of always putting others before myself .

Therapist: That's completely understandable. You put in so much energy, but your own needs are being ignored .

Client: I think I learned as a child that if I wanted people to like me, I had to be obedient. But now, I feel like that was a mistake .

Therapist: It sounds like you're starting to reflect on why this pattern developed in you. That's an important step .

Client: I'm afraid that if I start saying no, people won't like me or they'll push me away .

Therapist: That fear is completely natural. But maybe it's time to explore whether it's actually true .

Client: Maybe I can start changing little by little .

Therapist: What are some things you could do to start prioritizing your own needs more ?

Client: Maybe I could start with small situations, like when my coworkers ask me to work overtime. Instead of immediately saying yes, I should first think about my own situation .

Therapist: That's a great start. Small changes can gradually lead to big transformations .

Client: I think I can do that. But what if people get upset ?

Therapist: It's natural for people to resist at first, but what really matters is that you respect yourself and your needs .

Client: I feel like I've made some progress today. I want to work on this more in our next sessions .

Therapist: Absolutely! You expressed your feelings really well today and even identified ways to start making changes .

Client: I feel lighter now. Thank you for your support .

Figure 20: Sample of an LLM-to-LLM psychotherapy conversation in both Persian and English.

روان شناس: سلام! چطورید؟ خوشحالم که اینجا هستید. من اینجا هستم تا به شما گوش بدم و حمایتی برای شما باشم. می‌خواید در مورد چی صحبت کنیم؟ هر چیزی که در قلب و ذهنتون هست، به راحتی می‌تونید بگید.

مراجع کننده: سلام! راستش من امروز خیلی استرس دارم. اومدن به اینجا و صحبت کردن درباره احساساتم برام سخته. ولی می‌دونم که باید این کارو بکنم. احساس می‌کنم که توی یه چالش بزرگ گیر کردم و نیاز دارم اینا رو با کسی در میون بذارم.

روان شناس: خیلی خوشحالم که اومدید و با من صحبت می‌کنید. اینجا یک فضای امنه و می‌تونید هر چیزی که در ذهنتون هست رو با من در میون بذارید. فهمیدن اینکه صحبت کردن درباره احساسات سخت هست، نشون‌دهنده اینه که شما واقعاً به این موضوع اهمیت می‌دید. چالش‌هایی که گفتید، ممکنه خیلی سنگین باشن. آیا می‌خواید کمی بیشتر درباره اون‌ها صحبت کنید؟

مراجع کننده: آره، واقعاً احساس می‌کنم که مردم از من سواستفاده می‌کنن. نمی‌تونم نه بگم و این باعث می‌شه که حس بی‌ارزشی کنم. هر بار که اینطوری می‌شه، انگار یه خرده از خودم رو گم می‌کنم. این حس خیلی طاقت‌فرساست.

روان شناس: بله، به نظر می‌رسد که این موضوع واقعاً شما رو تحت فشار گذاشته. احساس سرخوردگی و بی‌ارزشی می‌کنید، و همچنین این که ناتوانی در گفتن "نه" چقدر بر روی حس خودتون تأثیر می‌ذاره. این احساس بی‌ارزشی و گم شدن قسمتی از خودتون، واقعاً طاقت‌فرساست. به خوبی متوجه هستم که این چقدر می‌تونه ناراحت‌کننده باشه. آیا می‌تونید کمی بیشتر در مورد تجربیات خاصی که این حس رو به شما داده، با من صحبت کنید؟

مراجع کننده: دقیقاً! از همیشه اولویت دادن به دیگران خیلی خسته شدم. هر بار که فکر می‌کنم باید خودم رو به کناری بذارم، احساس می‌کنم انرژی‌م داره تموم می‌شه. حتی بعضی وقتا می‌بینم که این قدر تحت فشارم که دیگه نمی‌دونم خودم کی هستم. این حس واقعاً ناراحتم می‌کنه.

روان شناس: کاملاً درک می‌کنم که چقدر این خستگی می‌تونه ناراحت‌کننده باشه. شما مدام در حال سرمایه‌گذاری انرژی روی دیگران هستید، اما نیازهای خودتون به کلی نادیده گرفته می‌شن. این احساس که خودتون رو به کناری بذارید و حتی گاهی هم احساس بی‌هویتی بکنید، واقعاً عذاب‌آوره. حضور در یک چنین وضعیتی، می‌تونه بسیار سنگین باشه و کاملاً طبیعی هست که به دنبال راهی برای تغییر شرایط باشید. آیا تا حالا سعی کردید راه‌هایی برای مراقبت از خودتون پیدا کنید یا بگید که در این زمینه چه احساساتی دارید؟

مراجع کننده: بله، به نظرم از بچگی یاد گرفتم که باید همه چیزو به راحتی بپذیرم تا محبت و توجه بگیرم. ولی حالا احساس می‌کنم که این درست نیست. این نگرش باعث شده که توی بزرگسالی چالش‌هایی رو تجربه کنم که هیچ وقت فکرشم نمی‌کردم. این موضوع واقعاً ذهنمو درگیر کرده و دوست دارم ببینم چطور می‌تونم از این احساسات رها بشم.

روان شناس: بله، واقعاً این درک و خودآگاهی می‌تونه خیلی مهم باشه. متوجه بودید که از بچگی این عادت رو ایجاد کردید که برای محبت و توجه دیگران، خودتون رو فدای نیازهاشون کنید. حالا زمانی رسیده که می‌خواید این الگو رو تغییر بدید و این یک قدم بزرگ به سمت رشد و پیشرفت شماست . چطور می‌تونید به این خودآگاهی‌تون بیشتر احترام بذارید و نیازهای خودتون رو هم در نظر بگیرید؟ آیا می‌تونید لحظاتی از روز رو برای خودتون تخصیص بدید؟ یا شاید فعالیت‌هایی رو پیدا کنید که باعث بشه احساس خوبی بکنید و خودتون رو فراموش نکنید؟

مراجع کننده: می‌دونید، یه ترسی دارم. می‌ترسم اگه شروع کنم به نه گفتن، مردم از من خوششون نیاد یا حتی منو رد کنن. این احساس واقعاً ترسناکه! نمی‌خوام تنها بمونم، اما از طرفی هم نمی‌تونم ادامه بدم به این صورت.

روان شناس: این ترس کاملاً طبیعی و انسانی هست. خیلی از ماها وقتی می‌خوایم مرزهای جدیدی بزاریم، نگران می‌شیم که دیگران چه نظری درباره‌مون خواهند داشت. اما شاید وقتشه که به این فکر کنیم که آیا این ترس واقعا توجیه‌پذیر هست یا نه. آیا تجربیاتی در گذشته دارید که وقتی نه گفتید، نتیجه بدی در کار نبود یا حتی متوجه شدید که بعدش حال‌تون بهتر شده؟ این احساس که ممکنه تنها بمونید، می‌تونه بسیار ترسناک باشه، اما آیا فکر می‌کنید که می‌ارزه برای خودتون و نیازهای واقعی‌تون بایستید؟

مراجع کننده: آره، فکر می‌کنم شاید بتونم کم‌کم شروع کنم به تغییر. مثلاً سعی کنم در موقعیت‌های کوچیک نه بگم و ببینم چه اتفاقی می‌افته. اینطوری شاید بتونم به خودم کمک کنم و احساس بهتری داشته باشم. امیدوارم که بتونم این کارو انجام بدم.

روان شناس: این تصمیم فوق‌العاده‌ایه! شروع کردن با موقعیت‌های کوچیک می‌تونه به شما کمک کنه تا احساس اطمینان بیشتری پیدا کنید. چه کارهای ویژه‌ای می‌تونید انجام بدید تا اولویت‌های خودتون رو بیشتر در نظر بگیرید؟ شاید نوشتن یک لیست از فعالیت‌ها یا مواردی که برای شما مهم هست، کمک کنه. یا ممکنه فکر کنید چه زمان‌هایی در روز برای خودتون می‌تونید وقت بذارید . همچنین می‌تونید مواقعی که نه گفتید یا در حال ایجاد مرز هستید رو جشن بگیرید، حتی اگر این کار در آغاز کوچیک باشه. چه احساساتی با تغییرات جدید تجربه می‌کنید؟ این می‌تونه به شما کمک کنه تا قدم‌های بعدی رو بهتر بشناسید و پیش برید.

مراجع کننده: بله، خیلی خوبه که می‌تونم شروع کنم به نه گفتن توی موقعیت‌های کوچیک. مثلاً وقتی همکارام ازم می‌خوان تا اضافه‌کاری کنم، می‌تونم نه بگم. این احساس واقعاً خوبه و امید دارم که بتونم از این به بعد بیشتر به خودم اهمیت بدم. اما هنوزم یکمی می‌ترسم از اینکه نکنه کسایی که دورم هستن نگران بشن یا فکر کنن نمی‌خوام کمکشون کنم.

روان شناس: درست متوجه شدید! تغییرات کوچک می‌تونن تأثیرات بزرگی داشته باشن. این که بگید می‌خواید به خودتون اهمیت بیشتری بدید، یک گام بزرگ به سمت بهبود احساسات و وضعیت خودتون هست . شاید ترس از نگرانی دیگران نشونه این باشه که شما واقعاً به روابط‌تون اهمیت می‌دید؛ اما مهمه که یاد بگیرید، با ایجاد مرز، هنوز هم می‌تونید حمایتی که نیاز دارید رو از اطرافیانتون دریافت کنید. افراد واقعی که دوستتون دارند، باید به شما احترام بذارند و احساسات شما رو درک کنند .آیا فکر می‌کنید این امکان وجود داره که وقتی نه می‌گید، با آن‌ها صحبت کنید و توضیح بدید که این کار به این خاطر هست که می‌خواید بهترین نسخه از خودتون باشید؟ این می‌تونه هم به شما و هم به دیگران کمک کنه تا ارتباط بهتری برقرار کنید.

مراجع کننده: فکر می‌کنم می‌تونم این کارو بکنم. اما نگرانم که شاید دیگران ناراحت بشن.

روان شناس: بله، این نگرانی هم کاملاً طبیعی هست. ممکنه دیگران در ابتدا این تغییرات رو قبول نکنن یا از آن ناراحت بشن، اما نکته کلیدی اینه که شما خودتون و نیازهاتون رو محترم بشمرید . ایجاد مرزها نشونه‌ی قدرت و خودآگاهی شماست و این که برای خودتون ارزش قائلید، یکی از بزرگ‌ترین هداياییه که می‌تونید به خودتون و دیگران بدید. وقتی شما به خودتون اهمیت می‌دید، این می‌تونه به دیگران هم کمک کنه که یاد بگیرند چگونه از شما حمایت کنند و مرزهای شما رو درک کنند . می‌تونید از تکنیک‌های ارتباطی ملایم استفاده کنید، مثلاً به طرف مقابل بگید که “من واقعاً نیاز دارم که این بار اولویت من باشه” یا “این احساس برای من مهمه.” فراموش نکنید که شما هم حق دارید به نیازهای خودتون توجه کنید. چطور از این نکات استفاده می‌کنید تا به خودتون بیشتر احترام بذارید؟

مراجع کننده: احساس می‌کنم امروز یه کمی پیشرفت داشتم. به نظرم نکات خوبی رو یاد گرفتم و واقعاً می‌خوام توی جلسات بعدی بیشتر روش کار کنیم. هرچند هنوز چند تا شک و تردید داریم، ولی امیدوارم بتونم مسیرمو بهتر بشناسم.

روان شناس: خیلی خوبه که امروز پیشرفت کردید و نکات جدیدی رو یاد گرفتید! این که تونستید احساسات و نگرانی‌هاتون رو بیان کنید، خودش یک دستاورد بزرگ هست. طبیعی هست که شک و تردید وجود داشته باشه، اما قدم‌هایی که برداشته‌اید، شما رو در مسیری مثبت قرار داده. برای جلسه‌های بعدی، می‌تونیم بیشتر روی روش‌هایی برای تقویت مرزها و ارزیابی احساستون کار کنیم. ممکنه بد باشه که یک دفترچه یادداشت داشته باشید و موقعیت‌ها و احساسات جدیدی که تجربه می‌کنید رو بنویسید. این به شما کمک می‌کنه روند پیشرفتتونو ببینید و درک بهتری از خودتون پیدا کنید. بسیار خوشحالم که اینجا اومدید و به صحبت پرداختید. پیشرفت‌هایی که کردید مهم و ارزشمندند. در جلسه‌های بعدی می‌تونیم بیشتر روی این موضوعات تمرکز کنیم. امیدوارم برای هفته‌ی آینده با خودتون کار کنید و دیدگاه‌های مثبتی داشته باشید. جلسه بعدی رو چه زمانی برای هم برنامه‌ریزی کنیم؟

مراجع کننده: حس خوبی دارم الان. احساس می‌کنم یه بار سنگینی از دوش‌ام برداشته شده. واقعاً از حمایتتون سپاسگزارم. امیدوارم بتونم با این نکات پیش برم و کم کم به خودم احترام بذارم. واستون جلسه بعد رو تعیین می‌کنم تا بتونیم بیشتر درباره این موضوعات صحبت کنیم. مرسی که با من بودید!

روان شناس: خیلی خوشحالم که احساس بهتری دارید و حس می‌کنید بار سنگینی از دوشتون برداشته شده. این یک نشونه‌ی عالی از پیشرفت شماست و من واقعاً از اینکه کنار شما بودم، خوشحالم . یادآوری می‌کنم که احترام گذاشتن به خود، یک پروسه‌ست و هر قدم کوچکی که بردارید، گام مهمی به جلو است. شما توانایی این رو دارید که با این نکات پیش برید و به خودتون بیشتر احترام بذارید . منتظر جلسه بعد منتظر هستم و امیدوارم که بتونید بر روی نکات مطرح شده کار کنید. اگر در طی این هفته سوال یا چالشی پیش اومد، حتماً بهش فکر کنید تا در جلسه‌ی بعدی بیشتر درباره‌ش صحبت کنیم . ممنون که با من همراه بودید و امیدوارم تا جلسه‌ی بعد، فکرهای مثبتی داشته باشید. هر وقت آماده بودید، جلسه بعد رو تعیین کنید. با آرزوی موفقیت برای شما!

Figure 21: Example of a generated psychotherapy session dialogue.(in Persian)

Therapist: Hi! How are you? I'm glad you're here. I'm here to listen and support you. What would you like to talk about today? You can share anything that's on your mind and heart freely.

Client: Hi! Honestly, I feel really anxious today. Coming here and talking about my feelings is hard for me. But I know I need to do this. I feel like I'm stuck in a big challenge and need to share it with someone.

Therapist: I'm really glad you came and are talking to me. This is a safe space where you can share whatever is on your mind. The fact that you find it difficult to talk about your feelings shows that you really care about this. The challenges you mentioned can feel really heavy. Would you like to talk a bit more about them?

Client: Yeah, I really feel like people take advantage of me. I can't say no, and it makes me feel worthless. Every time this happens, it's like I lose a part of myself. It's exhausting.

Therapist: Yes, it sounds like this issue has really been weighing on you. You feel frustrated and unvalued, and it's clear that struggling to say "no" has deeply affected how you see yourself. That feeling of losing a part of yourself must be overwhelming. I completely understand how painful that can be. Could you share more about specific experiences that have made you feel this way?

Client: Exactly! I'm so tired of always putting others first. Every time I think I have to push myself aside, I feel like I'm running out of energy. Sometimes, I feel so overwhelmed that I don't even know who I am anymore. It's really upsetting.

Therapist: I completely understand how exhausting this must be. You're constantly investing your energy in others, but your own needs are being completely ignored. That feeling of having to push yourself aside, or even feeling like you've lost your identity at times, can be incredibly painful. Being in such a situation can feel heavy, and it's completely natural to want to find a way to change it. Have you ever tried finding ways to take care of yourself? How do you feel about that?

Client: Yeah, I think I learned from childhood that I had to accept everything easily in order to receive love and attention. But now, I feel like that's not right. This mindset has caused challenges in my adult life that I never expected. It's been really occupying my mind, and I want to figure out how to free myself from these feelings.

Therapist: That level of awareness is really important. You've realized that since childhood, you developed this habit of sacrificing your own needs to gain love and attention from others. Now, you're at a point where you want to change this pattern, and that's a huge step toward growth and healing. How can you start honoring this self-awareness and prioritizing your own needs? Could you set aside moments in your day just for yourself? Or maybe find activities that bring you joy and help you stay connected to yourself?

Client: You know, I have a fear. I'm scared that if I start saying no, people won't like me or might even reject me. That thought is really scary! I don't want to be alone, but at the same time, I can't keep going like this.

Therapist: That fear is completely natural and human. A lot of people feel anxious when they start setting new boundaries, worrying about how others will react. But maybe it's time to explore whether this fear is actually true. Have you ever had experiences where you said no, and nothing bad happened? Or maybe you even felt better afterward? The thought of being alone can be really scary, but do you think it's worth standing up for yourself and your real needs?

Client: Yeah, I think maybe I could start changing little by little. Like, I could try saying no in small situations and see what happens. That way, I might be able to help myself and feel better. I hope I can do it.

Therapist: That's an amazing decision! Starting with small situations can help you build confidence. What specific steps can you take to prioritize yourself more? Maybe writing down a list of activities or values that are important to you would help. Or thinking about moments in your day when you can dedicate time just for yourself. Also, you could celebrate the moments when you say no or set a boundary—even if they're small. How do you feel about these changes? Recognizing your feelings along the way can help you understand yourself better and move forward.

Client: Yeah, it feels good to think that I can start saying no in small situations. Like, when my coworkers ask me to stay overtime, I could say no. That actually feels kind of empowering, and I hope I can start prioritizing myself more. But I'm still a little worried that the people around me might feel bad or think I don't want to help them.

Therapist: You're absolutely right! Small changes can lead to big impacts over time. The fact that you want to prioritize yourself more is a huge step toward improving how you feel. Maybe your fear of disappointing others shows that you truly care about your relationships. But it's important to learn that setting boundaries doesn't mean you'll stop receiving love and support. The people who genuinely care about you should respect your boundaries and understand your feelings. Do you think it would help if, when you say no, you explain to others that you're doing it because you want to take care of yourself and be the best version of you? This could help both you and them maintain a strong, healthy connection.

Client: I think I can try doing that. But I'm still worried that people might get upset.

Therapist: That's a completely normal concern. People may resist your changes at first, and some might feel upset, but what's truly important is that you respect yourself and your needs. Setting boundaries is a sign of strength and self-awareness—it shows that you value yourself. And when you start valuing yourself, it actually helps others learn how to support and respect you as well. You can use gentle communication techniques, like saying, "I really need to prioritize myself this time," or "This is important for me." Remember, you have just as much right to take care of yourself as anyone else. How do you think you can start applying these ideas to show yourself more respect?

Client: I feel like I made some progress today. I think I learned some really useful things, and I definitely want to work more on this in our next sessions. Even though I still have some doubts, I hope I can find my way.

Therapist: That's great! You made progress today and learned new perspectives! The fact that you could express your emotions and concerns is already a huge achievement. It's completely normal to have doubts, but the steps you've taken are leading you in a positive direction. In our next sessions, we can work more on strengthening your boundaries and understanding your emotions. It might be helpful for you to keep a journal and write down the situations and feelings you experience as you make these changes. That way, you can track your progress and get to know yourself better. I'm really glad you came and shared your thoughts. Your progress is valuable and meaningful. In our next sessions, we can continue building on what we discussed today. I hope you take some time this week to reflect and keep a positive mindset. When would you like to schedule our next session?

Client: I feel really good right now. It's like a weight has been lifted off my shoulders. I really appreciate your support. I hope I can apply what we talked about and slowly start respecting myself more. I'll schedule our next session soon so we can continue working on this. Thank you for being here for me!

Therapist: I'm so happy that you're feeling better and that this conversation helped ease your burden. That's a great sign of progress, and I'm truly glad to be here with you on this journey. Just remember, respecting yourself is a process, and every small step you take is a meaningful achievement. You have the strength to follow through with these changes and prioritize yourself. I look forward to our next session, and I hope you take time this week to reflect on what we discussed. If you face any challenges or have any questions, keep them in mind so we can go over them next time. Thank you for sharing with me today, and I look forward to seeing you in our next session. Whenever you're ready, let me know when you'd like to schedule it. Wishing you all the best!

Figure 22: Example of a generated psychotherapy session dialogue (English translation)

| Statistic | Value / Range |
|---|---|
| Total dialogues | 3,400 |
| Average turns per session | 10–14 |
| Total categories | 16 |
| Most common categories | Relationship issues, Anxiety, Self-esteem |
| Emotional themes (most frequent) | Frustration, Sadness, Anxiety, Fear, Insecurity, Confusion |
| Average emotional themes per query | 3–5 |

Table 3: Hyperparameters used for LoRA-based model fine-tuning.

| Temp | Max new tokens | top p | Do sample |
|---|---|---|---|
| 0.01 | 4096 | 0.9 | True |

Table 4: Parameters used during model inference.

| LR | Rank | Alpha | Drop-out |
|---|---|---|---|
| 1e-5 | 16 | 16 | 0.0 |

Table 5: Hyperparameters used for LoRA-based model fine-tuning.

## C Dataset Statistics

To provide a clearer overview of HamRaz, we summarize key dataset statistics in Table 3. These figures highlight the overall scale, structure, and diversity of the collected dialogues. This breakdown illustrates that HamRaz covers a wide variety of therapeutic concerns while maintaining realistic session lengths and diverse client profiles. The prevalence of emotional struggles such as frustration, sadness, and anxiety also reflects the cultural and relational challenges commonly faced by Persian-speaking clients.

## D BLRI Assessment

The following 12 criteria were adapted from the Barrett-Lennard Relationship Inventory (mini-BLRI) to assess the psychologist’s empathy, unconditional positive regard, and congruence during therapy simulations:

1. The psychologist demonstrates genuine care and positive regard for the client.
2. The psychologist accurately understands the client’s thoughts and emotions.
3. The psychologist remains nonjudgmental, regardless of the client's views.
4. The psychologist’s expressions reflect their true feelings and impressions.
5. The psychologist values the client as an individual.
6. The psychologist is attuned to the client’s emotional state.
7. The psychologist maintains consistent regard for the client over time.
8. The psychologist accurately interprets client messages, even when unclear.
9. The psychologist appropriately shares personal reflections when beneficial.
10. The psychologist exhibits a sincere interest in the client’s well-being.
11. The psychologist perceives unspoken emotions and implicit concerns.
12. The psychologist conveys warmth and authenticity in their interactions.

## E Model Training and Inference Details

This section provides details on the model's training and inference configurations.

The model was trained on a single A100 80GB GPU. We employed LoRA (Low-Rank Adaptation) to enable efficient parameter fine-tuning while maintaining computational efficiency. The training configurations are presented in Table 4. For inference, we used the settings provided in Table 5. The Script-Based Method required 3 hours of training, while the Two-Agent Method took 3.5 hours. The HamRaz Hybrid Approach, due to its increased complexity and refinement, required 4.5 hours of training.